# Neural Dynamics of AI-attributed Irony Reveal a Partial Intentional Stance

Xiaohui Rao[1], Hanlin Wu[1], Zhenguang G. Cai[1, 2,*]

[1] Department of Linguistics and Modern Languages [2] Brain and Mind Institute
The Chinese University of Hong Kong Hong Kong SAR

## Author notes

Correspondence should be addressed to Zhenguang G. Cai, Department of Linguistics and Modern Languages, Leung Kau Kui Building, The Chinese University of Hong Kong, Shatin, N.T., Hong Kong; zhenguangcai@cuhk.edu.hk.

**Abstract**

As Large Language Models (LLMs) are increasingly deployed as social agents and trained to produce humor and irony, a question emerges: when encountering witty AI remarks, do people interpret them as deliberate communicative acts? This study investigated whether people adopt an *intentional stance*, ascribing mental states to explain behavior, when comprehending AI-attributed irony. Irony provides a testbed because understanding it requires distinguishing intentional contradictions from unintended errors through pragmatic reanalysis. Using electroencephalography (EEG) to measure event-related potentials (ERPs), we compared behavioral and neural responses to identical ironic utterances attributed to either an AI companion or a human. We found that people do not fully adopt an intentional stance towards AI communication. Participants interpreted contextually incongruent utterances as irony significantly less often when attributed to an AI than when attributed to a human. Correspondingly, we observed attenuated neural responses to AI-attributed irony compared to human-attributed irony during both initial semantic processing (P200) and pragmatic reanalysis (P600). In addition, these neural responses were modulated by individual perceptions of AI sincerity and trustworthiness, with positive perceptions facilitating pragmatic comprehension by reducing cognitive effort. This suggests that adopting an intentional stance toward AI is a flexible and adaptive process, dynamically shaped by people's mental models of artificial agents. These findings reveal that despite advances in communicative competence, AI systems face a barrier to social agency: humans process input from artificial interlocutors with reduced ascription of intentionality. This has important implications for understanding human social cognition and for designing AI systems intended for social interaction.

## Introduction

Large Language Models (LLMs) have transformed AI from functional tools into social agents serving as virtual tutors[1,2], mental health chatbots[3], trip advisors[4], and AI friends[5]. Achieving social agency hinges on their ability to handle complex, human-like interaction. As these systems are increasingly trained to use humor and irony for social rapport building[6,7], a question emerges: do people ascribe communicative intentions to witty remarks from an AI companion? In daily life, people navigate their social interaction by explaining and predicting others' behavior with reference to others' mental states; this cognitive strategy is known as adopting an *intentional stance*[8]. For example, when someone says "You are always so punctual!" to a late arrival, we assume the speaker intends irony, rather than making a factual error. This intentional stance is a basic step of Theory of Mind, the cognitive capability to understand that others have different perspectives[9,10], and is critical for non-literal language comprehension, successful communication, and empathic understanding[11]. Given the emerging social roles of LLMs, a crucial question is whether people extend this intentional stance to AI companions.

### *From human-computer to human-AI language interaction*

According to the *Computers Are Social Actors* framework[12], people tend to apply human social scripts (i.e., the ingrained social rules and expectations that guide human-to-human interaction) to computers, responding with politeness[13], reciprocity[14], and demographic stereotypes[15]. For instance, in a study by Nass and colleagues[13], participants interacted with a computer and were subsequently asked to evaluate its performance. The results showed that participants rated its performance more positively when the same computer asked them for feedback, compared to when a different computer (located in a separate room) or a paper-and-pencil questionnaire asked for the evaluation[13]. This suggests people treat computers politely, avoiding direct criticism similar to how people behave during human-to-human interaction[13,16,17]. This automatic application of interpersonal norms to computers may occur as a default social reflex, independent of whether people actually ascribe conscious, human-like qualities to the machine[12,18]. This automatic application of human scripts also permeates more complex linguistic interactions. For instance, people rated irony from an early, pre-LLM AI system as comparably ironic and high in intentionality as human-generated irony[19]. As these early systems lacked social agency, these high ratings likely reflect people mindlessly applying human social scripts, projecting their own interpretive frames or treating the machine as an extension of the human programmer's agency, rather than inferring a mental state within the machine itself[12,18,19].

Nevertheless, when interacting with artificial agents with sufficient agency and human-like characteristics (e.g., robots), people tend to shift from automatic social ascription to deliberate mental state inference[20], driven by the need to understand and predict a robot's complex behavior[18]. Neurophysiological evidence supports this by showing that brain activation associated with Theory of Mind correlates positively with a robot's human-likeness[20]. Recent research further reveals that people often explain robot behavior through intentional terms (e.g., desires and beliefs) rather than mechanistic terms (e.g., programming and hardware)[21]. Intentional stance adoption increases when robots exhibit social behaviors such as eye gaze, gestures, and emotional expressions, or possess human-like appearance[22]. While intentional stance adoption revealed by these studies relies primarily on embodied cues of robots, contemporary text-based LLMs interact with humans almost solely through their human-like language performance[23]. Recent neuroimaging evidence suggests that the intentional stance can be triggered by language alone, provided people are actively inferring the identity of their interlocutor[24]. While this evaluative goal recruits the brain's mentalizing network[25,26], the authors' analyses of neural activity within this network reveal a behavioral-neural dissociation[24]. Even when people failed to behaviorally distinguish a chatbot from a human in subjective ratings, machine-learning classifiers can still decode the agent's artificiality from multi-voxel activity patterns in mentalizing regions like the DMPFC and rTPJ[24–26]. While humans may consciously treat competent AI language as intentional, they implicitly register its artificial origin at the neural level.

Beyond neural activation, this sensitivity to AI identity alters the real-time mechanics of language comprehension through a process known as interlocutor modeling[27]. In human communication, people construct a model of their interlocutors based on important demographics such as age, gender, or social status to adjust their own language output[28,29] and to constrain their language comprehension[27,30–34]. Recent EEG evidence reveals distinct processing strategies for LLM-attributed language compared to human-attributed language, reflecting unconscious adaptation to perceived LLM characteristics. For instance, comprehenders exhibited attenuated N400 responses to semantic anomalies attributed to an LLM compared to a human[35], indicating reduced cognitive effort required for semantic integration[36]. Conversely, LLM-attributed syntactic errors elicited an enhanced P600 than human-attributed errors[35], signaling an intensified structural reanalysis[37]. Rather than processing linguistic input passively, comprehenders actively construct a mental model of the LLM (e.g., anticipating factual unreliability while expecting grammatical precision) and comprehend its language against this model[35]. This neural adaptation also extends to social-communicative contexts. For instance, when comprehending humorous puns, people exhibited smaller N400 responses to LLM-attributed jokes than to human-attributed ones[38], reflecting reduced initial cognitive effort to resolve the semantic incongruity[36]. Conversely, LLM-attributed jokes elicited an enhanced Late Positive Potential (LPP) during the later stages of comprehension[38]. Together, these neural signatures indicate a unique processing strategy: people might initially expect less semantic depth from the LLM, but undergo intensified affective and cognitive updating when forced to reconcile those low expectations with the LLM's surprisingly complex delivery[36,38–40]. However, these studies examined neural responses to semantic, syntactic, and humorous content rather than communicative intentions. Investigating intentional stance adoption towards AI is essential, as inferring communicative goals is important to treating an LLM-powered AI companion as a communicative partner. Crucially, irony comprehension provides a testbed for examining intentional stance adoption because it involves pragmatic reanalysis to distinguish a deliberate communicative intent from an unintended semantic error[19,41]. Therefore, how people process AI-attributed irony can reveal whether they perceive these AI systems as intentional communicative partners.

### *The present study*

The primary aim of the present study was to test the hypothesis that humans attribute an intentional stance to artificial agents. To investigate this, we examined both behavioral (i.e., an irony recognition and categorization task) and neural (i.e., event-related potentials; ERPs) responses to ironic versus literal utterances attributed to an AI companion versus a human. This approach allows us to track both explicit comprehension and the underlying neurophysiological processes involved in social cognition, as EEG's millisecond temporal resolution is essential for revealing the cognitive paths taken by comprehenders in real time[42]. Specifically, we manipulated the perceived identity of the interlocutor (AI vs. Human) and the text participants read. In particular, participants were told that they were to read a mini-dialogue between two humans or between a human and an AI companion. The mini-dialogue consisted of a statement (e.g., "我今天吃了一天的炸鸡", meaning "I ate fried chicken all day today") and a response (e.g., "你吃得真健康啊！", meaning "You eat so healthily"). We measured participants' electrophysiological responses (i.e., ERPs) during the reading of the response. We tested participants' intentional stance via a post-experiment irony recognition and categorization task: Participants first judged whether the response's literal meaning was congruent with the preceding statement. Then, for responses judged as incongruent, they categorized the response as an ironic response ("irony"), as an erroneous response resulting from misunderstanding the statement ("misunderstanding"), as a polite response ("politeness"), or as a response due to other reasons ("others").

To track the cognitive dynamics of intentional stance adoption in real time, we focused on established ERP components that index different stages of irony processing. The earliest stage is usually tracked by the P200, an early positive deflection emerging around 200 ms post-stimulus that reflects initial incongruity detection and early semantic processing[43–48]. Crucially, this early component can be modulated by top-down pragmatic knowledge about the interlocutor. For instance, the P200 amplitude is larger for sentences that match a speaker's specific communicative style (such as irony spoken by a typically ironic speaker, or a literal statement spoken by a non-ironic speaker)[46]. While the N400, a centro-parietal negativity peaking around 400 ms post-stimulus, is often associated with lexical retrieval or integration[36], this component is typically absent during successful irony comprehension with strong contextual support, as the context facilitates pragmatic resolution[43–46]. Exploratorily, a domain-general "oddball" P300 to ironic statements may emerge in this mid-latency window, potentially reflecting the reallocation of attentional resources toward the unexpected ironic stimulus[43,45,49,50]. Finally, a late positive deflection emerges around 500 ms after stimulus onset, known as the P600. While associated with structural repair[37,51] or thematic anomalies[52,53], in the context of irony, the P600 has been suggested to reflect the cognitive effort required to reject a literal interpretation and impute a deliberate, non-literal intent[43–46,54]. Because this late-stage reanalysis is highly sensitive to interlocutor context[46,55–57], the P600 serves as a critical neural index in our study for intentional stance adoption, allowing us to track whether comprehenders attribute a deliberate communicative goal to an AI's linguistic incongruity.

Depending on the extent to which comprehenders extend this intentional stance to AI, we expected our behavioral and neural outcomes to fall along a continuum of three plausible predictions. If participants adopt a *full* intentional stance, we would expect them to treat AI irony equivalently to human irony, yielding comparable proportions of "irony" categorizations and similar P600 effects (i.e., difference wave amplitudes between the ironic and literal conditions), indicating successful pragmatic reanalysis. If participants adopt a *partial* intentional stance towards AI, we would anticipate a reduced ascription of deliberate intent to the AI, reflected in lower "irony" categorizations and attenuated P600 amplitudes. Finally, if people adopt *no* intentional stance, treating AI semantic incongruities as unintended errors, we would expect

"misunderstanding" to be the dominant behavioral categorization, potentially accompanied by enhanced N400 effects reflecting semantic violation rather than pragmatic reanalysis[36].

A secondary aim was to explore the modulatory role of individual perceptions of AI sincerity and trustworthiness. Our focus on perceived AI sincerity and trustworthiness was guided by research that relies on these constructs to evaluate human-AI interaction[38,58–60]. Recent ERP evidence reveals that these perceptions modulate the processing of AI-attributed puns. Specifically, research indicates that people who perceive an AI as more sincere and trustworthy exhibit reduced N400 amplitudes (facilitating semantic integration), while those who perceive an AI as more trustworthy show enhanced LPP amplitudes (reflecting intensified interlocutor model updating)[38]. Building on these initial findings, one possible outcome was that positive perceptions of AI would reduce the overall effort required to resolve pragmatic ambiguity. However, because sincerity (a social-affective trait) and trustworthiness (a cognitive-evaluative trait)[61] may tap into different aspects of human-AI interaction, we left the temporal dynamics of these effects as an open question.

## Methods

### *Design*

We used a task where participants read a mini-dialogue consisting of a statement by a person and a response which was either attributed to another person or to an AI companion. Attribution of the response was manipulated between participants and within items: participants were told that they would read mini-dialogues either between two humans or between a human and an AI companion, following a previous paradigm[58]. We manipulated the statement (within participants and within items) such that the response would be best interpreted ironically (i.e., in the ironic condition; e.g., "I ate fried chicken all day today" followed by "You ate so healthily") or literally (i.e., in the literal condition; "I ate salad all day today" followed by "You ate so healthily"). Thus, the experiment had a mixed factorial design of 2 (Attribution: AI vs. human) x 2 (Irony: ironic vs. literal).

### *Participants*

Sixty-four neurologically healthy native Mandarin speakers were recruited (36 females, 28 males; mean age 23 years old, range 18-35). Five were excluded (two for excessive artifacts, two for equipment malfunction, and one for failing the manipulation check), yielding a final sample of 59 participants (34 females, 25 males; mean age = 23.02, range = 18-35; see Data recording and preprocessing for details). Among the final sample, 30 participants were assigned to the Human condition and 29 to the AI condition. Independent samples *t*-tests indicated no significant difference in age between the Human and AI conditions ($t(57) = -0.498$, $p = 0.620$). A Chi-square test indicated no significant difference in gender distribution between the two conditions ($\chi^2(1) = 0.81$, $p = 0.367$). All participants provided written informed consent and received monetary compensation (HK$180). The study received approval from the Joint Chinese University of Hong Kong-New Territories East Cluster Clinical Research Ethics Committee. The study was not preregistered.

### *Materials*

There were 80 sets of written mini-dialogues in Chinese, developed on the basis of similar stimuli used in previous studies[62,63]. Each mini-dialogue set refers to a pair of mini-dialogues that share the same target response but differ in their preceding context statement (one biasing a literal interpretation, the other biasing an ironic interpretation). Each mini-dialogue within these sets consisted of a statement produced by a person and a response attributed to a human interlocutor or an AI companion (see Table 1). The statement was in the form of a first-person declarative sentence describing a personal behavior, situation or experience (e.g., “我今天吃了一天的炸鸡”, “I ate fried chicken all day today”). The response was in the form of subject + an intensifier (真, “so”) + a critical adjective that conveys a subjective evaluation (e.g., “你吃得真健康啊”, “You ate so healthily”). The critical words were familiar Chinese adjectives, varying between one and two characters in length. The literal or ironic nature of the response was manipulated by varying the preceding statement to be either semantically congruent or incongruent with the evaluative adjective in the response. In the literal condition, the statement described a situation that matched the adjective (e.g., eating salad followed by a praise of healthiness). In the ironic condition, the statement described a situation that contradicted the adjective (e.g., eating fried chicken followed by a praise of healthiness), thereby triggering an ironic interpretation through semantic violation.

We conducted two norming tests on the stimuli with independent samples of native Chinese speakers who did not participate in the main EEG experiment. First, 20 participants evaluated their subjective familiarity of the critical words on a 7-point Likert scale (1 = very unfamiliar, 7 = very familiar). The critical stimuli obtained a high mean familiarity score of 6.70 (*SD* = 0.25, range = 5.85-6.95), confirming that these stimuli were very familiar to the participants. Second, a separate group of 20 participants rated the degree of irony for all mini-dialogues on a 7-point Likert scale (1 = not at all ironic, 7 = very ironic). A paired-samples *t*-test confirmed that responses in the ironic condition were perceived as significantly more ironic ($M = 6.59$, $SD = 0.22$), than those in the literal condition ($M = 1.59$, $SD = 0.32$; $t(78) = 127$, $p < .001$).

Following a Latin Square design, we constructed two stimulus lists, each containing 40 ironic and 40 literal mini-dialogues, with each mini-dialogue set appearing in only one condition per list. Each list also included 80 filler mini-dialogues featuring neutral, non-evaluative exchanges to minimize response bias. For instance, a filler mini-dialogue involved one interlocutor saying, “最近我工作太忙了，没时间去旅行” (“I’m too busy with work recently and don’t have time to travel”), and the other responding, “你可以试着周末出游”(“You can try traveling on the weekend”).

**Table 1**. The example of the stimuli.

| Irony | Statement (Interlocutor 1) | Response (Interlocutor 2) |
|---|---|---|
| Ironic | 我 今天 吃了 一天 的 炸鸡。 | 你 吃得 真 **健康** 啊！ |
| | *wo1 jin1tian1 chi1le yi4tian1 de zha2ji1* | *ni3 chi1de zhen1 jian4kang1 a* |
| | “I ate fried chicken all day today.” | “You ate so **healthily**!” |
| Literal | 我 今天 吃了 一天 的 沙拉。 | 你 吃得 真 **健康** 啊！” |
| | *wo1 jin1tian1 chi1le yi4tian1 de sha1la1* | *ni3 chi1de zhen1 jian4kang1 a* |
| | “I ate salad all day today.” | “You ate so **healthily**!” |

The original Chinese text shown to the participants (top), the Pinyin romanization (middle), and the English translation (bottom). Bolded text indicates the critical word.

***Procedure***

**Questionnaires.** Participants’ general perceptions of LLM-powered AI as an interlocutor were assessed using 7-point semantic differential scales adapted from previous research[38,58–60]. In this method, participants rated AI along a continuum between two polar adjectives (e.g., 1 = dishonest, 7 = honest), with 4 representing a neutral midpoint. Trustworthiness was measured using three bipolar adjective pairs: dishonest-honest, untrustworthy-trustworthy, and unreliable-reliable. Sincerity was evaluated with three additional pairs: insincere-sincere, artificial-genuine, and performed-heartfelt. To capture baseline perceptions, participants in the AI condition completed these measures prior to the main experiment. Conversely, participants in the human condition completed the measures post-experiment to minimize the possibility that prior exposure to AI-related questions would raise suspicion and undermine their belief in the human interlocutor.

**The main experiment.** A Wizard-of-Oz methodology was employed to create a controlled environment where participants’ beliefs about their interlocutor could be manipulated[64]. Participants were led to believe they would read transcribed mini-dialogues between two humans or between a human and their AI companion, when in fact, all mini-dialogues were pre-programmed and controlled by the experimenter using E-Prime 3.0. To prevent participants from mapping prior expectations onto specific commercial models (e.g., GPT-4), we introduced the AI as a custom, lab-developed system to standardize baseline assumptions. Each trial began with a central fixation cross displayed for 1000 ms, followed by the presentation of the statement (Interlocutor 1) for 5000 ms. After a brief 500 ms inter-stimulus interval, a second central fixation cross appeared for 1000 ms. Subsequently, the response (Interlocutor 2) was presented segment-by-segment. Each word appeared for 700 ms, with an inter-segment interval of 200 ms. The adjective word in the response served as the critical word for the EEG recording. To ensure engagement and comprehension, a judgment task was presented on half of both the target trials and filler trials. This task probed the understanding of the statement content. For example, following a mini-dialogue between 小莉 ("Xiaoli") and her AI companion, participants were presented with a probe statement like "请判断：小莉在谈论自己吃了什么。" ("Please judge: Xiaoli is discussing what she ate"). Participants were required to judge whether the probe statement was congruent with the preceding dialogue by pressing one of two keyboard keys (e.g., J/F).

**Manipulation check for source attribution.** To validate the experimental manipulation, following previous studies[65], at the end of the EEG experiment, we had participants report who they thought had produced the responses in the mini-dialogues (Interlocutor 2). Of the 64 participants, 63 correctly reported the Interlocutor 2 as either a human or an AI companion, consistent with their assigned condition. One participant in the human condition reported disbelief that the mini-dialogues involved two humans and was therefore excluded from subsequent analyses.

**Irony recognition and categorization task.** Following the main experiment, participants completed an irony recognition and categorization task which was aimed at assessing how participants interpreted the response. As shown in Figure 1, participants were presented with all target trials from the main experiment. For each trial, the full text of the mini-dialogue was presented in consecutive lines. They first judged whether the response's literal meaning was congruent with the preceding statement. For responses judged as incongruent, participants then categorized the response as an ironic response ("irony"), as an erroneous response resulting from misunderstanding the statement ("misunderstanding"), as a polite response ("politeness"), or as a response due to other reasons ("others"). Selecting "irony" indicated recognition of deliberate communicative intent (intentional stance), whereas selecting "misunderstanding" indicated categorizing the incongruity as an unintended error. For responses categorized as "irony", participants made a subsequent forced-choice judgment of the ironic motivation as either "sarcastic" or "humorous". This secondary classification was included because verbal irony serves a complex set of social and communicative goals, ranging from affiliative humor to aggressive mockery and criticism[66]. Exploring this allowed us to test whether the identity of the interlocutor (AI vs. Human) modulated the perceived pragmatic function of the ironic statement.

**Figure 1**. Flowchart of the irony recognition and categorization task.

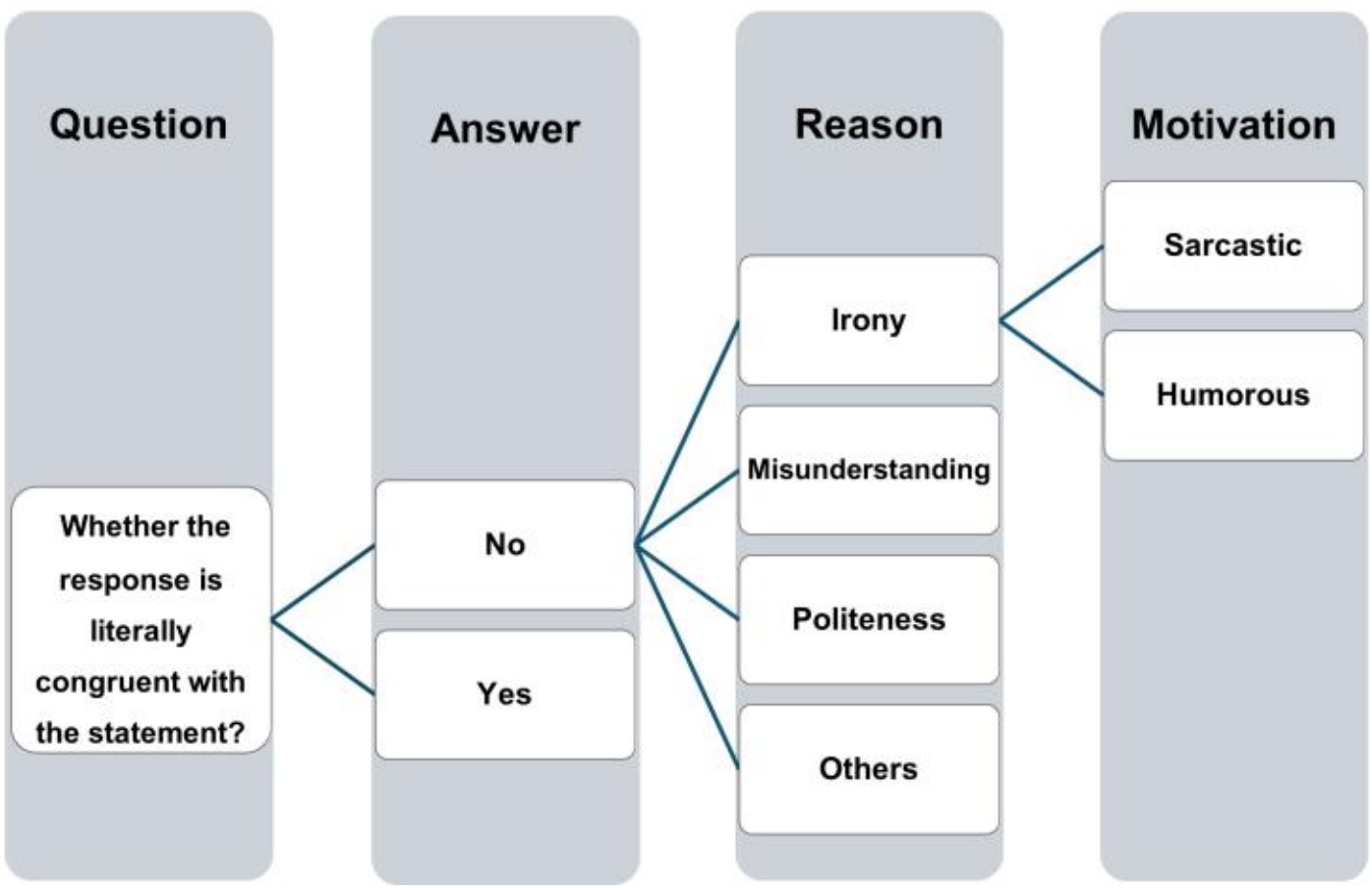

**Figure 2**. Study experimental design.

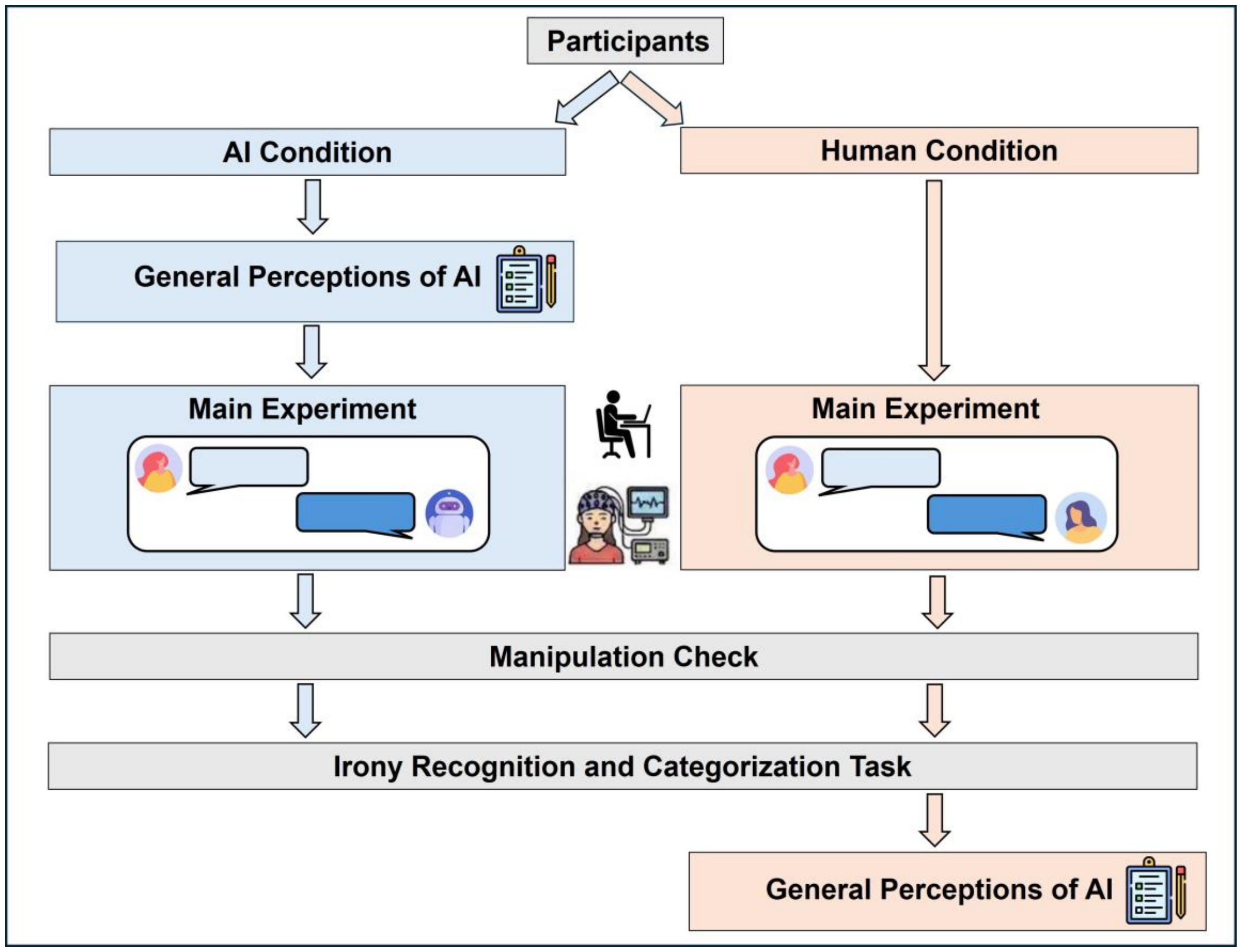

Flowchart of the experimental procedure. Participants were assigned to an AI or a human condition, completed the main interaction task, followed by a manipulation check and an irony recognition and categorization task. General perceptions of AI were collected at the beginning (AI condition) and at the end (human condition) of the study.

### *Data recording and preprocessing*

EEG activity was recorded using a 128-channel g.HIamp amplifier with active Ag/AgCl electrodes. Electrodes were positioned according to the extended 10-20 system and referenced online to the left earlobe. Data were sampled at 512 Hz, with electrode impedances maintained below 30 kΩ. Preprocessing was conducted using custom MATLAB scripts and the FieldTrip toolbox[67]. The continuous EEG was bandpass-filtered between 0.1 and 30 Hz and re-referenced to the average of both earlobes[68]. To remove ocular artifacts from the high-density recordings, independent component analysis was performed on the filtered data with dimensionality reduced to 30 components[69,70]. The data were segmented into epochs spanning -200 ms to 900 ms relative to the critical word (i.e., adjective) onset, with baseline correction applied using the -200 ms to 0 ms pre-stimulus interval. Epochs containing voltage deflections exceeding ±150 μV were rejected, resulting in 6.17% trial loss. Five participants were excluded: two for excessive artifact contamination (>30% rejected trials), two for equipment/procedure malfunction, and one for failing the manipulation check, yielding a final sample of 59 participants (34 females, 25 males; mean age 23.02 years old, range 18-35). Accuracy rates on the judgment task of the main experiment were high for the remaining participants ($M$ = 94.49%, $SD$ = 4.47%, range = 80-100%), indicating they were attentive and engaged with the mini-dialogues in the main experiment. We further removed 3.31% of the trials from the EEG analysis on the basis of the post-experiment irony recognition and categorization task: a trial was removed if the participant judged the response's literal meaning as congruent with the preceding statement in the ironic condition or as incongruent in the literal condition. Statistical analyses were conducted using Generalized Linear Mixed-effects (GLME) models for behavioral data and Linear Mixed-effects (LME) models for ERP data. We verified that all models satisfied the necessary statistical assumptions through model-specific residual diagnostics: for LME models, we verified the normality of residuals and homogeneity of variance using standard visual inspection (Q-Q plots and residual-versus-fitted plots); for GLME models, we verified that the simulated residuals were uniformly distributed using DHARMa analysis. The R analysis scripts used for these diagnostics are available in our OSF repository.

## Results

### *Behavioral results*

As shown in Figure 3, the majority of responses in the ironic condition were categorized as ironic responses ("irony"). To examine how participants categorized the incongruent responses across AI and human conditions, we analyzed the proportions of "irony",

“misunderstanding”, and “politeness” using three separate GLME models at the trial level. Each model used binary coding for the target category (e.g., “irony” = 1, others = 0) and included Attribution (AI vs. Human) as the fixed main effect, with Participant and Item as random effects (Table 2). For all GLME analyses, we used forward model comparison (with $\alpha = 0.2$) to identify the maximal random-effects structure supported by the data[71]. As shown in Table 2, the GLME analyses revealed that participants were significantly less likely to categorize AI-attributed incongruities as ironic responses ($N = 634$) compared to human-attributed ones ($N = 842$; $\beta = -1.52$, 95% CI [-2.57, -0.47], $SE = 0.54$, $z = -2.83$, $p = .005$). Conversely, participants were significantly more likely to categorize AI incongruities as erroneous responses resulting from misunderstanding the statement ($N = 455$) compared to human ones ($N = 177$; $\beta = 2.89$ [1.56, 4.23], $SE = 0.68$, $z = 4.24$, $p < .001$). Participants also showed a marginally significant tendency to interpret fewer AI-attributed incongruities as polite responses ($N = 40$) compared to human-attributed ones ($N = 102$; $\beta = -1.57$ [-3.19, 0.05], $SE = 0.83$, $z = -1.90$, $p = .058$). Finally, a GLME on responses categorized as ironic showed no significant difference in sarcastic versus humorous motivations between the AI and human conditions ($\beta = 0.09$ [-0.64, 0.83], $SE = 0.37$, $z = 0.24$, $p = .807$).

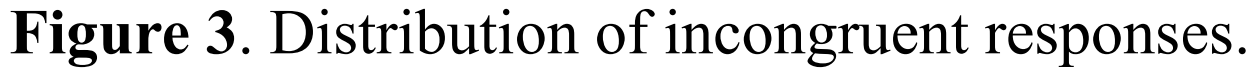

**Figure 3**. Distribution of incongruent responses.

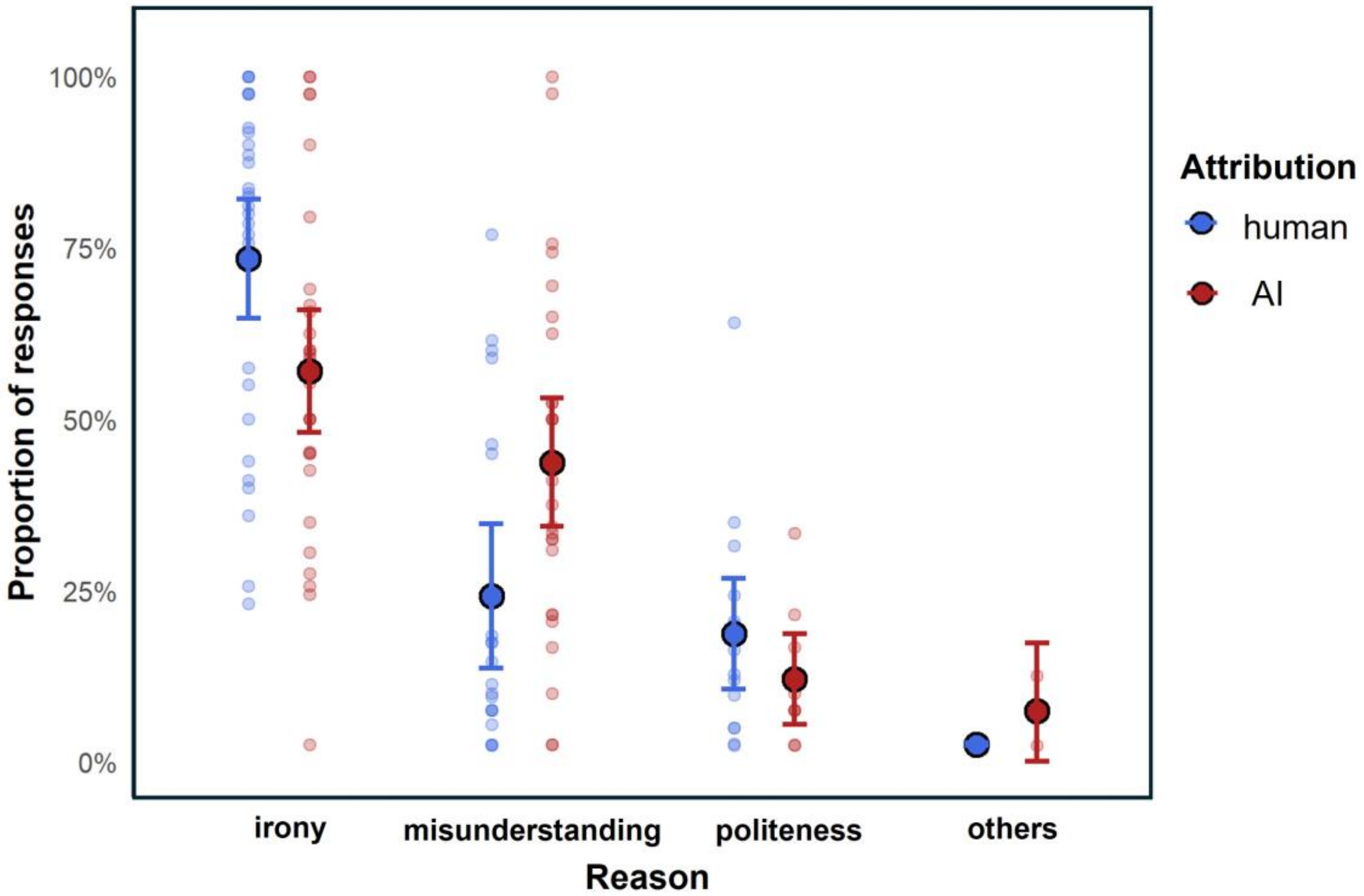

The distribution of reasons for incongruent responses across Attribution (human vs AI). Solid points represent the group-level mean of participant proportions for each category. Error bars represent 95% confidence intervals of the mean. Semi-transparent jittered points display the underlying distribution of individual participant-level proportions (*N* = 59 participants; *N* = 30 in the human condition, *N* = 29 in the AI condition).

**Table 2**. GLME models for behavioral analyses.

| *Fixed effects* | *β [95% CI]* | *SE* | *z* | *p* |
|---|---|---|---|---|
| Irony analysis | | | | |
| Intercept | 1.19 [0.63, 1.75] | 0.29 | 4.14 | < .001 |
| Attribution | -1.52 [-2.57, -0.47] | 0.54 | -2.83 | .005 |
| | | | | |
| Misunderstanding analysis | | | | |
| Intercept | -2.09 [-2.79, -1.38] | 0.36 | -5.82 | < .001 |
| Attribution | 2.89 [1.56, 4.23] | 0.68 | 4.24 | < .001 |
| | | | | |
| Politeness analysis | | | | |
| Intercept | -5.03 [-6.17, -3.89] | 0.58 | -8.64 | < .001 |
| Attribution | -1.57 [-3.19, 0.05] | 0.83 | -1.90 | .058 |

Model for irony analysis: Irony ~ Attribution + (1 | Participant) + (1 | Item); Model for misunderstanding analysis: Misunderstanding ~ Attribution + (1 | Participant) + (Attribution + 1 | Item); Model for politeness analysis: Politeness ~ Attribution + (1 | Participant) + (1 | Item).

***EEG results***

To prevent analytical bias and control the familywise error rate in time-window selection, we adopted a uniform, data-driven collapsed localizer approach prior to conducting any condition-level statistical comparisons[68,72]. Following this established methodology, we generated a single grand-averaged waveform collapsed across all participants and all experimental conditions across the full cap. Because this grand average remains blind to condition-specific variances, visual inspection of its morphology allows for the objective identification of latent component boundaries without inflating the Type I error rate[72].

As illustrated by the grand-averaged waveform (see Figure S1 in the Supplementary Materials), distinct deflections and peaks emerged across the epoch. Based on these data-driven boundaries, we defined three contiguous analysis windows. Two of these captured our hypothesized components: 150-300 ms to capture the early post-sensory P200 component (reflecting initial incongruity detection), and 450-900 ms to encompass the late positive P600 component indexing pragmatic reanalysis or meaning integration[43–46,54]. The objective localizer also revealed an intermediate 300-450 ms window. While this mid-latency range is typically evaluated for classic negative-going N400 effects indexing lexical-semantic integration difficulties[36], our collapsed grand average prominently exhibited a positive-going deflection. This aligns with literature that under strong contextual support, a traditional irony-related N400 may be absent[43–46], and a domain-general "oddball" P300 may emerge instead due to the unexpected/rare nature of ironic statements[45,49,50]. The ERP waveforms across experimental conditions and corresponding scalp topographies for each of these three analytical epochs are illustrated in Figure 4.

To investigate the topographical distribution of the reported effects on ERP amplitudes, we conducted posteriority and laterality analyses[33,73]. Scalp sites were divided into four quadrants: left-anterior, right-anterior, left-posterior, and right-posterior, with each region comprising 17 channels (see Table S1 in the Supplementary Materials). For the posteriority analysis, separate LME models were fit across the three time windows with Irony, Attribution, and Posteriority (posterior vs anterior) entered as interacting fixed effects, alongside Word Length as a covariate (see Table S2 for full model structures). The three-way interaction of Irony × Attribution × Posteriority did not reach statistical significance in any of the analyzed epochs (150-300 ms: $\beta$ = 0.02 [-0.81, 0.86], *SE* = 0.43, $t(16645.46) = 0.06$, $p = .954$; 300-450 ms: $\beta$ = 0.13 [-0.91, 1.17], *SE* = 0.53, $t(16580.81) = 0.25$, $p = .801$; 450-900 ms: $\beta$ = 0.11 [-1.06, 1.28], *SE* = 0.60, $t(16576.82) = 0.19$, $p = .854$). Furthermore, the two-way interaction between Irony and Posteriority was not significant (150-300 ms: $\beta$ = -0.15 [-0.57, 0.27], *SE* = 0.21, $t(16645.46) = -0.70$, $p = .482$; 300-450 ms: $\beta$ = -0.13 [-0.65, 0.39], *SE* = 0.27, $t(16580.81) = -0.50$, $p = .617$; 450-900 ms: $\beta$ = 0.31 [-0.28, 0.90], *SE* = 0.30, $t(16577.42) = 1.04$, $p = .300$). Similarly, a laterality analysis was performed using separate LME models with Irony, Attribution, and Laterality (left vs right) entered as interacting fixed effects, and Word Length included as a covariate. Across all the time windows, neither the three-way interaction of Irony × Attribution × Laterality reached significance (150-300 ms: $\beta$ = -0.02 [-0.86, 0.82], *SE* = 0.43 $t(16660.42) = -0.05$, $p = .960$; 300-450 ms: $\beta$ = -0.12 [-1.16, 0.92], *SE* = 0.53, $t(16580.61) = -0.23$, $p = .820$); 450-900 ms: $\beta$ = -0.14 [-1.32, 1.04], *SE* = 0.60, $t(16655.20) = -0.24$, $p = .814$) nor did the two-way

interaction of Irony × Laterality (150-300 ms: $\beta = 0.09$ [-0.33, 0.51], $SE = 0.21$, $t(16660.40) = 0.40$, $p = .689$; 300-450 ms: $\beta = -0.11$ [-0.63, 0.41], $SE = 0.27$, $t(16580.61) = -0.43$, $p = .669$; 450-900 ms: $\beta = -0.11$ [-0.70, 0.48], $SE = 0.30$, $t(16655.20) = -0.35$, $p = .724$). These topographical analyses indicated that both the main effect of Irony and the modulatory effect of Attribution on irony processing were broadly distributed across the scalp, as evidenced by the absence of significant topographical interactions across the three time windows.

Consequently, to maximize statistical power while eliminating arbitrary electrode selection bias, we focused our primary amplitude analyses on an electrode cluster of 62 central channels covering frontal, central, and parietal sites for all three time windows (with grand-averaged ERP waveforms and a spatial visualization of these 62 channels provided in Figure 4). This approach captured broad ERP modulation over the entire scalp, while excluding peripheral boundary channels that are prone to muscle and eye-movement noise[74]. Rather than collapsing these electrodes into a single regional average, we retained individual channel data, treating Channel as a random effect in our LME models. This approach accounts for baseline voltage differences across the scalp while preserving unique spatial variance, a statistical framework utilized in recent ERP literature[35,75,76]. As implementing multiple LMEs with multiple fixed factors across sequential epochs can inflate the experimentwise and familywise error rates[77], we corrected the $p$-values of all critical fixed effects from the LME models using the False Discovery Rate (FDR) method across all time windows. Where significant interaction effects were detected within a specific time window, follow-up separate analyses were conducted, with the resulting p-values controlled via FDR corrections separately for that window, following previous correction protocols[78,79].

**Figure 4**. Neural responses to ironic and literal statements.

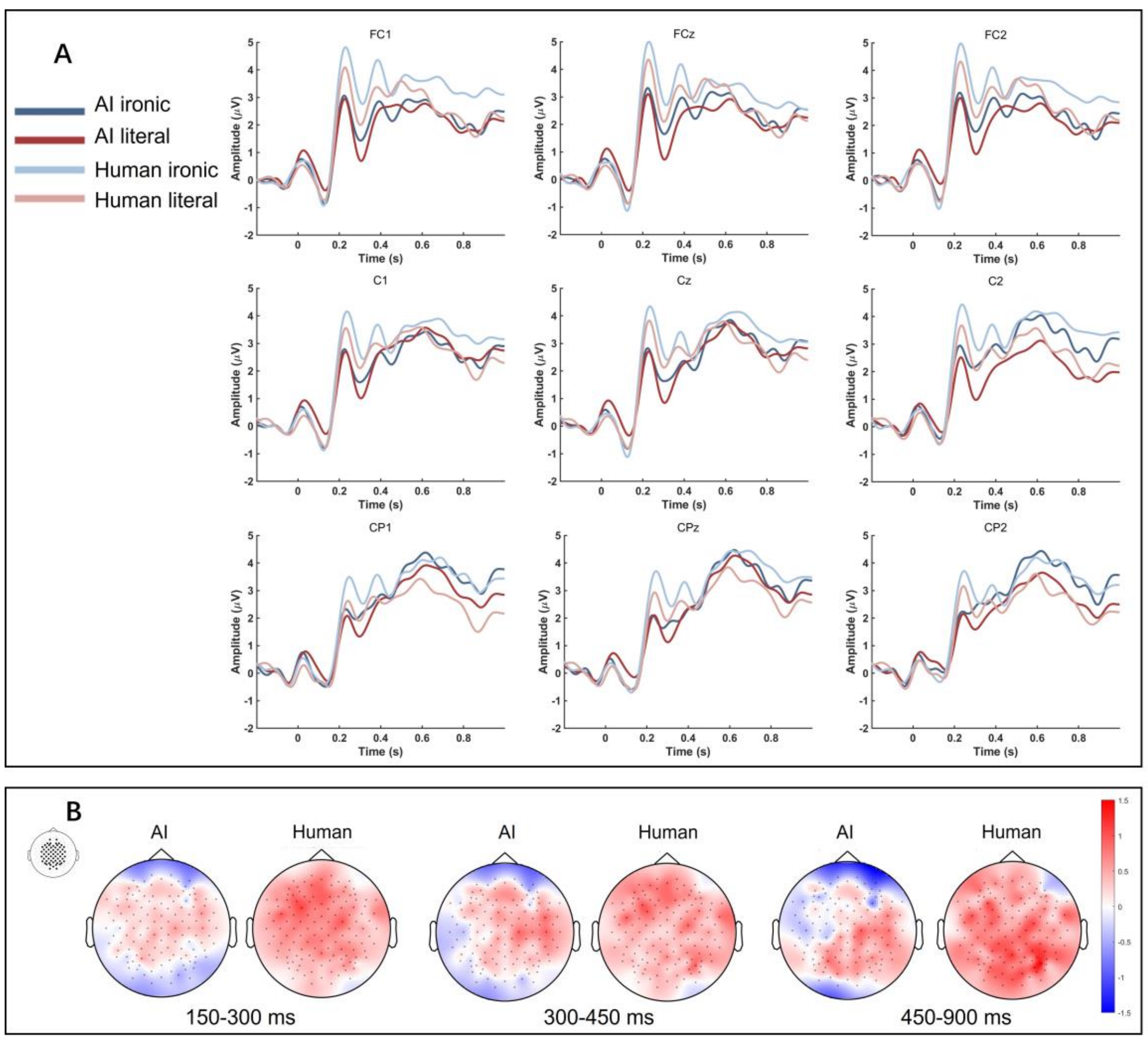

(A) Amplitudes of ironic and literal statements in the AI and human conditions; (B) Topographies of the Irony effects (Ironic minus Control) during 150-300 ms, 300-450 ms, and 450-900 ms after critical word onset, and spatial visualization of the electrode cluster of 62 central channels for amplitude analyses (see Table S1). All data presented in this figure are derived from $N = 59$ participants ($N = 30$ in the human condition, $N = 29$ in the AI condition).

We primarily analyzed neural responses using LME models with Irony (ironic vs. literal) and Attribution (AI vs. human) as interacting fixed effects, while controlling for Word Length as a fixed-effect covariate across three time windows. Participant, Item, and Channel were included as random effects (see Table 3). Forward model comparison was used for all LME analyses (with $\alpha = 0.2$). As shown in Figure 4 and Table 3, analysis of the 150-300 ms window revealed a main effect of Irony ($\beta$ = 0.38 [0.31, 0.44], *SE* = 0.03, $t$(543.22) =11.42, $p$ < .001, *FDR-corrected p* < .001), with a more positive P200 for ironic relative to literal statements. Although the uncorrected model suggested a main effect of Attribution ($\beta$ = -0.83 [-1.63, -0.04], *SE* = 0.41, $t$(57.46) = -2.05, $p$ = .045), this effect did not survive FDR correction (*FDR-corrected p* = .134). Crucially, there was an interaction between Irony and Attribution ($\beta$ = -0.39 [-0.51, -0.26], *SE* = 0.06, $t$(263597.03) = -5.99, $p$ < .001, *FDR-corrected p* < .001), indicating that the P200 effect (Ironic minus Literal) was smaller in the AI condition (0.21 μV; $\beta$ = 0.18 [0.09, 0.28], *SE* = 0.05, $t$(2532.18) = 3.86, $p$ < .001, *FDR-corrected p* < .001) than in the human condition (0.50 μV; $\beta$ = 0.58 [0.49, 0.66], *SE* = 0.05, $t$(439.19) = 12.80, $p$ < .001, *FDR-corrected p* < .001). During 300-450 ms, a main effect of Irony was found ($\beta$ = 0.38 [0.31, 0.46], *SE* = 0.04, $t$(263478.51) =9.53, $p$ < .001, *FDR-corrected p* < .001), showing a more positive P300 for ironic statements than literal statements. The interaction between Irony and Attribution was not significant ($\beta$ = 0.04 [-1.01, 1.08], *SE* = 0.53, $t$(77.78) =0.07, $p$ = .946), suggesting the P300 effect (Ironic minus Literal) was comparable across AI and human conditions. During 450-900 ms, a main effect of Irony was observed ($\beta$ = 0.51 [0.42, 0.61], *SE* = 0.05, $t$(367.71) = 10.49, $p$ < .001, *FDR-corrected p* < .001), revealing a more positive P600 for ironic relative to literal statements. An interaction between Irony and Attribution was observed ($\beta$ = -0.30 [-0.48, -0.12], *SE* = 0.09, $t$(263595.49) = -3.22, $p$ = .001, *FDR-corrected p* = .002), indicating that the P600 effect (Ironic minus Literal) was smaller in the AI condition (0.35 μV; $\beta$ = 0.35 [0.21, 0.49], *SE* = 0.07, $t$(334.93) = 4.82, $p$ < .001, *FDR-corrected p* < .001) than in the human condition (0.62 μV; $\beta$ = 0.69 [0.57, 0.81], *SE* = 0.06, $t$(3195.21) = 11.00, $p$ < .001, *FDR-corrected p* < .001).

**Table 3**. LME models for main amplitude analyses.

| *Fixed effects* | *β [95% CI]* | *SE* | *df* | *t* | *p* |
|---|---|---|---|---|---|
| P200 (150-300 ms) | | | | | |
| Intercept | 1.86 [1.38, 2.34] | 0.25 | 110.96 | 7.57 | < .001 |
| Irony | 0.38 [0.31, 0.44] | 0.03 | 543.22 | 11.42 | < .001 |
| Attribution | -0.83 [-1.63, -0.04] | 0.41 | 57.46 | -2.05 | .045 |
| Word Length | 0.38 [0.13, 0.63] | 0.13 | 77.95 | 2.99 | .004 |
| Irony : Attribution | -0.39 [-0.51, -0.26] | 0.06 | 263597.03 | -5.99 | < .001 |
| P300 (300-450 ms) | | | | | |
| Intercept | 2.56 [1.79, 3.33] | 0.39 | 83.05 | 6.52 | < .001 |
| Irony | 0.38 [0.31, 0.46] | 0.04 | 263478.51 | 9.53 | < .001 |
| Attribution | -0.47 [-1.95, 1.00] | 0.75 | 72.85 | -0.63 | .532 |
| Word Length | 0.22 [-0.11, 0.55] | 0.17 | 77.99 | 1.33 | .189 |
| Irony : Attribution | 0.04 [-1.01, 1.08] | 0.53 | 77.78 | 0.07 | .946 |
| P600 (450-900 ms) | | | | | |
| Intercept | 2.67 [1.97, 3.37] | 0.36 | 101.51 | 7.46 | < .001 |
| Irony | 0.51 [0.42, 0.61] | 0.05 | 367.71 | 10.49 | < .001 |
| Attribution | -0.07 [-1.25, 1.12] | 0.61 | 56.99 | -0.11 | .914 |
| Word Length | 0.63 [0.27, 0.98] | 0.18 | 78.00 | 3.46 | < .001 |
| Irony : Attribution | -0.30 [-0.48, -0.12] | 0.09 | 263595.49 | -3.22 | .001 |

Model for main amplitude analysis (150-300 ms): Amplitude ~ Irony * Attribution + Word Length + (1 | Participant) + (1 | Item) + (Irony + Attribution + 1 | Channel); Model for main amplitude analysis (300-450 ms): Amplitude ~ Irony * Attribution + Word Length + (1 | Participant) + (Attribution + Irony : Attribution + 1 | Item) + (1 | Channel); Model for main amplitude analysis (450-900 ms): Amplitude ~ Irony * Attribution + Word Length + (1 | Participant) + (1 | Item) + (Irony + 1 | Channel). *P*-values are uncorrected. For FDR-corrected values of critical experimental effects, see the Results section.

To examine whether neural responses to AI-attributed ironic versus literal utterances were modulated by participants' perceptions of AI sincerity and trustworthiness, we constructed separate LME models with Irony × Sincerity and Irony × Trustworthiness as interacting fixed effects across the three time windows for the AI condition, with Word Length included as a fixed-effect covariate and Participant, Item, and Channel as random effects (Tables 4 and 5).

During the 150-300 ms window, the sincerity analysis showed an interaction between Irony and Sincerity ($\beta$ = -0.32 [-0.41, -0.23], *SE* = 0.05, $t(131268.90)$ = -7.00, $p < .001$, *FDR-corrected* $p < .001$), indicating that the P200 effect in the AI condition decreased as a function of perceived AI sincerity (Figure 5). During 300-450 ms, only a main effect of Irony was observed ($\beta = 0.40$ [0.28, 0.52], $SE = 0.06$, $t(131933.77) = 6.63$, $p < .001$, *FDR-corrected* $p < .001$). During 450-900 ms, there was a main effect of Irony ($\beta = 0.36$ [0.22, 0.50], $SE = 0.07$, $t(338.04) = 4.98$, $p < .001$, *FDR-corrected* $p < .001$), and an interaction between Irony and Sincerity ($\beta = -0.17$ [-0.30, -0.03], $SE = 0.07$, $t(131069.48) = -2.44$, $p = .015$, *FDR-corrected* $p = .033$), indicating that the P600 effect in the AI condition decreased as a function of perceived AI sincerity (Figure 5).

During 150-300 ms, the trustworthiness analysis found no main effect of Irony ($\beta = 0.21$ [-0.46, 0.87], $SE = 0.34$, $t(23.56) = 0.61$, $p = .545$) or interaction between Irony and Trustworthiness ($\beta = -0.08$ [-0.82, 0.65], $SE = 0.37$, $t(23.35) = -0.22$, $p = .826$). However, during 300-450 ms, a main effect of Irony was found ($\beta = 0.43$ [0.31, 0.55], $SE = 0.06$, $t(131926.87) = 7.10$, $p < .001$, *FDR-corrected* $p < .001$). An interaction between Irony and Trustworthiness was also observed ($\beta = -0.33$ [-0.46, -0.20], $SE = 0.06$, $t(131903.17) = -5.12$, $p < .001$, *FDR-corrected* $p < .001$), indicating that P300 effect in the AI condition decreased as perceived AI trustworthiness increased (Figure 5). During 450-900 ms, there was a main effect of Irony ($\beta = 0.38$ [0.24, 0.52], $SE = 0.07$, $t(338.50) = 5.23$, $p < .001$, *FDR-corrected* $p < .001$) and an interaction between Irony and Trustworthiness ($\beta = -0.40$ [-0.55, -0.26], $SE = 0.07$, $t(131905.14) = -5.43$, $p < .001$, *FDR-corrected* $p < .001$), indicating that P600 effect in the AI condition decreased as perceived AI trustworthiness increased (Figure 5).

**Table 4**. LME models for amplitude analyses of sincerity in the AI condition.

| *Fixed effects* | *β [95% CI]* | *SE* | *df* | *t* | *p* |
|---|---|---|---|---|---|
| P200 (150-300 ms) | | | | | |
| Intercept | 1.47 [0.78, 2.15] | 0.35 | 46.23 | 4.21 | < .001 |
| Irony | 0.22 [-0.39, 0.84] | 0.31 | 78.06 | 0.71 | .481 |
| Sincerity | -0.03 [-0.61, 0.56] | 0.30 | 28.81 | -0.09 | .933 |
| Word Length | 0.47 [0.15, 0.79] | 0.16 | 77.39 | 2.91 | .005 |
| Irony : Sincerity | -0.32 [-0.41, -0.23] | 0.05 | 131268.90 | -7.00 | < .001 |
| | | | | | |
| P300 (300-450 ms) | | | | | |
| Intercept | 2.32 [1.11, 3.54] | 0.62 | 34.59 | 3.76 | .001 |
| Irony | 0.40 [0.28, 0.52] | 0.06 | 131933.77 | 6.63 | < .001 |
| Sincerity | 0.05 [-1.06, 1.15] | 0.56 | 27.98 | 0.08 | .934 |
| Word Length | 0.32 [-0.07, 0.71] | 0.20 | 78.00 | 1.59 | .115 |
| Irony : Sincerity | 0.03 [-0.08, 0.15] | 0.06 | 131082.02 | 0.57 | .571 |
| | | | | | |
| P600 (450-900 ms) | | | | | |
| Intercept | 2.67 [1.62, 3.72] | 0.54 | 42.15 | 4.97 | <.001 |
| Irony | 0.36 [0.22, 0.50] | 0.07 | 338.04 | 4.98 | <.001 |
| Sincerity | -0.30 [-1.20, 0.61] | 0.46 | 27.98 | -0.64 | .528 |
| Word Length | 0.86 [0.42, 1.31] | 0.23 | 77.97 | 3.83 | <.001 |
| Irony : Sincerity | -0.17 [-0.30, -0.03] | 0.07 | 131069.48 | -2.44 | .015 |

Model for sincerity analysis (150-300 ms): Amplitude ~ Irony * Sincerity + Word Length + (1 | Participant) + (Irony + 1 | Item) + (1 | Channel); Model for sincerity analysis (300-450 ms): Amplitude ~ Irony * Sincerity + Word Length + (1 | Participant) + (1 | Item) + (1 | Channel); Model for sincerity analysis (450-900 ms): Amplitude ~ Irony * Sincerity + Word Length + (1 | Participant) + (1 | Item) + (Irony + 1 | Channel). *P*-values are uncorrected. For FDR-corrected values of critical experimental effects, see the Results section.

**Table 5**. LME models for amplitude analyses of trustworthiness in the AI condition.

| *Fixed effects* | *β [95% CI]* | *SE* | *df* | *t* | *p* |
|---|---|---|---|---|---|
| P200 (150-300 ms) | | | | | |
| Intercept | 1.44 [0.79, 2.10] | 0.33 | 43.77 | 4.32 | < .001 |
| Irony | 0.21 [-0.46, 0.87] | 0.34 | 23.56 | 0.61 | .545 |
| Trustworthiness | 0.47 [-0.17, 1.11] | 0.33 | 27.95 | 1.43 | .163 |
| Word Length | 0.54 [0.25, 0.83] | 0.15 | 73.53 | 3.65 | < .001 |
| Irony : Trustworthiness | -0.08 [-0.82, 0.65] | 0.37 | 23.35 | -0.22 | .826 |
| P300 (300-450 ms) | | | | | |
| Intercept | 2.30 [1.10, 3.50] | 0.61 | 34.77 | 3.76 | .001 |
| Irony | 0.43 [0.31, 0.55] | 0.06 | 131926.87 | 7.10 | <.001 |
| Trustworthiness | 0.50 [-0.75, 1.76] | 0.64 | 27.98 | 0.79 | .438 |
| Word Length | 0.31 [-0.08, 0.70] | 0.20 | 78.04 | 1.58 | .119 |
| Irony: Trustworthiness | -0.33 [-0.46, -0.20] | 0.06 | 131903.17 | -5.12 | <.001 |
| P600 (450-900 ms) | | | | | |
| Intercept | 2.62 [1.58, 3.66] | 0.53 | 42.36 | 4.92 | <.001 |
| Irony | 0.38 [0.24, 0.52] | 0.07 | 338.50 | 5.23 | <.001 |
| Trustworthiness | 0.49 [-0.55, 1.52] | 0.53 | 28.00 | 0.92 | .365 |
| Word Length | 0.87 [0.42, 1.31] | 0.22 | 78.02 | 3.85 | <.001 |
| Irony : Trustworthiness | -0.40 [-0.55, -0.26] | 0.07 | 131905.14 | -5.43 | <.001 |

Model for trustworthiness analysis (150-300 ms): Amplitude ~ Irony * Trustworthiness+ Word Length + (Irony +1 | Participant) + (1 | Item) + (1 | Channel); Model for trustworthiness analysis (300-450 ms): Amplitude ~ Irony * Trustworthiness + Word Length + (1 | Participant) + (1 | Item) + (1 | Channel); Model for trustworthiness analysis (450-900 ms): Amplitude ~ Irony * Trustworthiness + Word Length + (1 | Participant) + (1 | Item) + (Irony +1 | Channel). *P*-values are uncorrected. For FDR-corrected values of critical experimental effects, see the Results section.

**Figure 5**. Modulation of irony processing by individual perceptions.

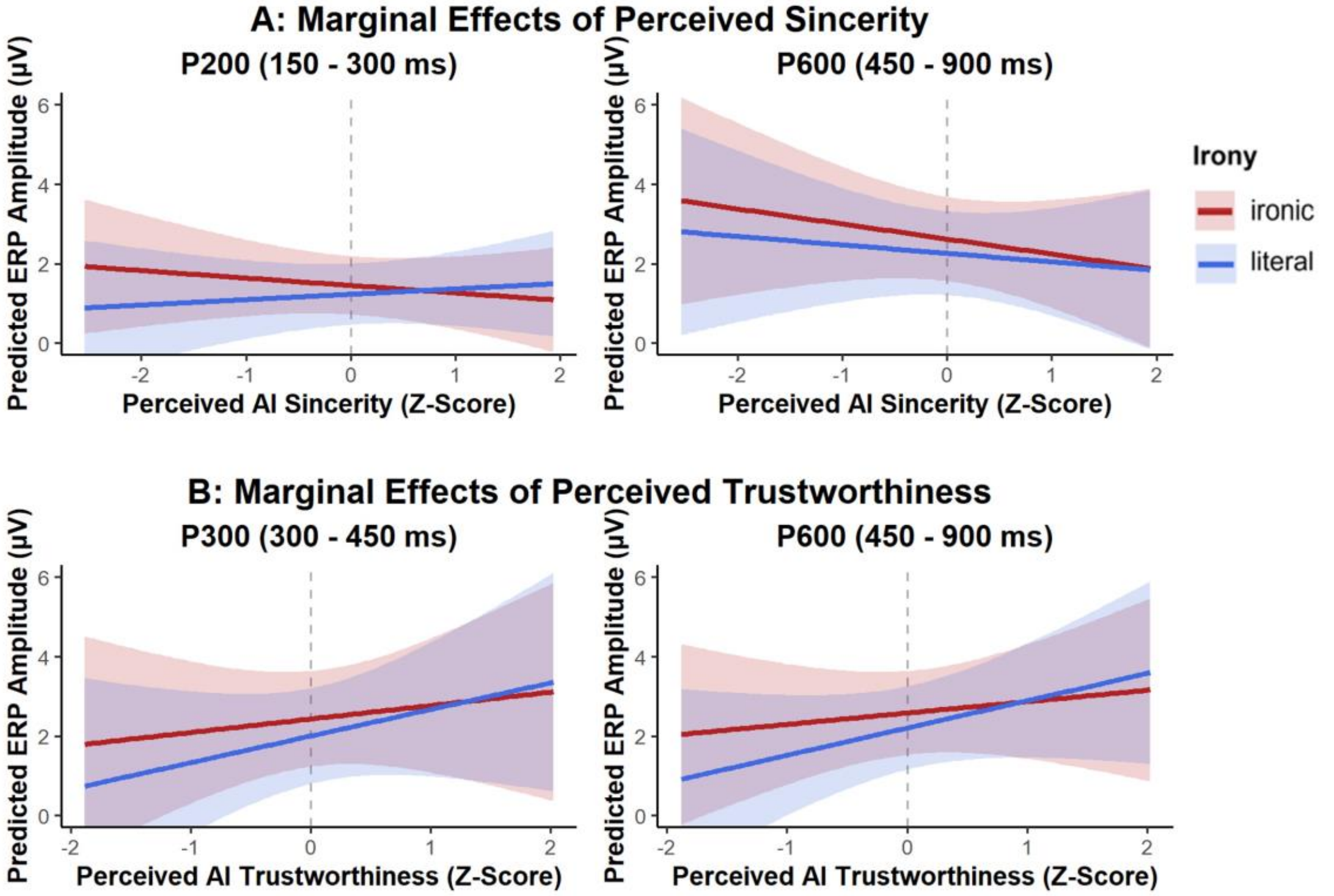

(A) Estimated marginal effects of perceived AI sincerity on the P200 (150-300 ms) and P600 (450-900 ms); (B) Estimated marginal effects of perceived AI trustworthiness on the P300 (300-450 ms) and P600 (450-900 ms). Solid lines represent model-predicted ERP amplitudes for ironic and literal conditions as a function of standardized ratings of perceived AI sincerity and trustworthiness. Shaded regions indicate 95% confidence intervals derived from the LME models ($N$ = 29 participants in the AI condition). The vertical dashed line at Z = 0 denotes the mean rating.

## Discussion

An ongoing area of research in cognitive science is whether our interaction with conversational AI relies on the same behavioral and neural mechanisms as human-to-human communication[80]. The present study provides a perspective to address this broader inquiry by investigating whether humans extend the intentional stance to AI companions. By utilizing EEG to track the real-time cognitive dynamics of irony comprehension, our behavioral and neural findings reveal a dissociation: while people can recognize irony from AI, they engage in different, less cognitively effortful processing than they do for human-attributed irony. This reflects the adoption of a *partial* intentional stance toward AI.

***Behavioral evidence for a partial intentional stance towards AI***

Our behavioral results showed that source attribution alters how people interpret utterances that are literally incongruent with the context. Irony remained the most common interpretation for incongruent utterances regardless of attribution condition. However, participants categorized fewer incongruent utterances as irony and more as errors resulting from misunderstanding when the source was attributed to be AI than to be human. This behavioral pattern challenges a strict interpretation of the *Computers Are Social Actors* framework[12]. While it posits that people automatically apply human social scripts to computers, our data suggests a boundary condition: this automaticity does not fully extend to the ascription of communicative intentionality. The larger proportion of "misunderstanding" judgments in the AI condition aligns with a mechanistic explanation[21], where semantic incongruity is interpreted as the AI's failure to understand the context rather than as a deliberate communicative act. This also suggests a boundary to anthropomorphism: while prior work has shown that people spontaneously ascribe human-like characteristics to AI, including age, gender, perspective-taking abilities, and linguistic competence[35,81–84], our results indicate that people still hesitate to attribute true agency and intentionality to these systems.

This pattern differs from earlier findings showing that people rated irony from primitive AI systems as comparably ironic and intentional as irony from a human[19]. While this discrepancy could be due to several methodological factors (e.g., differences in the language of the stimuli, experimental design, or the specific types of irony used), we suggest that this pattern may reflect a shift in how people process increasingly sophisticated AI. Simple AI systems elicited automatic and mindless social ascription, but more advanced and human-like LLMs appear to trigger deliberate mental state evaluation, consistent with findings that greater human-likeness of robots prompts effortful intentional stance adoption[20]. Recent theoretical models characterize these modern LLMs as “minimal cognitive agents”[85] that simulate intentionality, a phenomenon termed “pseudo-consciousness”[86], without possessing subjective experience. Our behavioral findings provide empirical evidence that humans recognize this ontological distinction, adopting a partial intentional stance that treats these agents as communicative partners for structural efficiency while selectively withholding the ascription of deep intentionality[85].

However, this skepticism appears to act primarily as an initial interpretive threshold. Among utterances categorized as ironic, perceived motivation (i.e., whether the irony served as affiliative humor or face-threatening sarcasm) did not differ by attribution. This suggests that the identity of interlocutors (e.g., AI vs. human) may primarily affect the initial inference of intentionality; once a deliberate intent is ascribed, it does not seem to alter the further interpretation of its specific social function. This dissociation is consistent with hierarchical models of social cognition[10,87]. Specifically, such frameworks conceptualize Theory of Mind as a two-stage process: the initial detection of an agent capable of intentionality, followed by the evaluation of that agent’s social goals[87]. Our findings indicate that while people adopt a partial intentional stance towards artificial agents, the subsequent evaluation of those mental states may rely on the recycling of standard, domain-general human social schemas[12]. In the absence of an established AI-specific cognitive template for complex pragmatics, comprehenders may default to deploying their existing human-to-human interpretive tools once the threshold of intentionality has been crossed.

### ***The neurocognitive signature of a partial intentional stance towards AI***

While our behavioral data show that source attribution drives the outcome of irony comprehension, our ERP findings reveal that this is not a static, post-hoc judgment. Rather, it is a dynamic neurocognitive process that adjusts to the attributed interlocutor across three distinct stages. First, AI-attributed irony elicited attenuated P200 effects compared to human-attributed irony. Given that the P200 has been associated with initial incongruity detection and semantic processing[46], this attenuation may suggest that people initially allocate less attention to AI-attributed semantic incongruity, possibly reflecting lower expectations of semantic reliability. While previous research has observed attenuated N400 effects for AI-attributed semantic violations and puns[35,38], our P200 data extends this reduction in cognitive effort to the social-pragmatic domain. This indicates that comprehenders lower their expectations for an AI's semantic reliability much earlier in the comprehension process. This early modulation aligns with predictive processing models of language[88], which posit that comprehenders use top-down models of interlocutor identity to rapidly pre-activate expected linguistic input. Our data suggests that the AI label acts as a top-down cue that down-weights semantic predictions before the words are fully integrated.

Second, the P300 component, linked to domain-general updating of existing mental models when encountering unexpected information[49,50], showed comparable effects for both AI-attributed and human-attributed irony. Because ironic sentences made up only 25% of the stimuli in our study, this low probability might trigger a general task-related "oddball" surprise response across both conditions[43,45,49,50]. The presence of this comparable P300 suggests that participants' cognitive engagement was comparable across both the AI and human conditions, investing similar cognitive effort to integrate the surprising information.

Third, AI-attributed irony elicited reduced P600 effects compared to human-attributed irony. Since the P600 has been associated with pragmatic reanalysis required to reinterpret incongruity as deliberate irony[45,54] and is known to be sensitive to interlocutor context[46,55–57], this attenuation may reflect diminished intentional stance adoption with fewer resources invested in recovering intended meaning from AI-attributed utterances. This finding adds a complementary dimension to recent literature. Prior work has shown that AI-attributed syntactic errors elicit heightened P600 effects compared to human errors, possibly because humans have higher expectations of AI's syntactic capabilities[35]. By shifting to complex pragmatics, we observe an attenuated P600, revealing a domain-specific modulation of the intentional stance.

Taken together, these three components reveal a dissociation: while domain-general processes of updating mental models (P300) remained constant across attribution conditions, both domain-specific early semantic processing (P200) and late pragmatic reanalysis (P600) were modulated by source attribution, supporting our hypothesis of partial intentional stance adoption (H2). These results provide neurophysiological evidence that the account of interlocutor modelling, in which language comprehension adjusts to speaker demographics such as age, gender, or social status[27,30–34], can be extended by incorporating an agent's perceived ontological status and the level of attributed intentionality. Notably, this real-time neural dynamic provides an update to recent fMRI literature. While prior studies argue that language from text-based AI only engages social-cognitive networks during explicit identity evaluation[24], our paradigm shows that the intent evaluation inherent to irony comprehension is sufficient to elicit a partial intentional stance, with the AI label alone serving as a sufficient cue for comprehenders to track the agent's artificiality.

Moreover, the adoption of this intentional stance is not uniform; it is modulated by personal beliefs about AI capabilities. Our findings support the hypothesis that positive perceptions facilitate pragmatic comprehension by reducing cognitive effort. Specifically, as participants perceived the AI as more sincere, the P200 and P600 effects for AI-attributed irony significantly decreased. This suggests that attributing a social-affective trait like sincerity[61] to an artificial agent reduces the initial semantic friction (P200) and lessens the neurocognitive burden of late-stage pragmatic reanalysis (P600). Similarly, higher AI trustworthiness attenuated both the P300 and P600 effects. Trust, a cognitive-evaluative trait[61], appears to facilitate the domain-general updating of unexpected information (P300) while also reducing the integration effort required to recover the AI's intended meaning (P600). Consistent with theories of human-automation interaction, trust acts as a cognitive shortcut, allowing users to rely on technology even when the system is too complex to fully monitor or understand[89]. In our findings, this manifests as a reduction in neurocognitive monitoring: interpersonal trust in conversational AI lowers the cognitive effort required to resolve pragmatic ambiguity. While a partial intentional stance represents the default processing mode when interacting with AI interlocutors, these continuous variables suggest that human neurocognitive processing strategies toward AI are malleable. This dynamic flexibility challenges the *Computers Are Social Actors* framework's reliance on static human-human social scripts[12] and updates recent arguments for specialized "human-media" scripts[90]. We provide neurocognitive evidence for this shift, indicating that the human mind does not treat AI as a monolithic category; rather, it maintains a malleable, real-time mental model of agency, continuously fine-tuned by individual perceptions and the perceived social identity of the interlocutor.

### *Limitations and future directions*

Our findings should be interpreted in light of several methodological constraints and opportunities for future research. While we controlled for baseline assumptions by introducing the AI as a custom, lab-developed AI, we did not measure prior familiarity with AI systems or perceptions of the AI's "smartness". Future studies should integrate these metrics, as people highly familiar with advanced LLMs may attribute implied meaning differently than those accustomed to more primitive systems. Second, there is a limitation regarding stimulus artificiality. The conversational dialogues were constructed for this experiment rather than drawn from spontaneous human-human or actual human-AI interaction. This reliance on pre-scripted text raises an open question: it is possible that participants perceived some utterances as somewhat unnatural, prompting them to rate incongruities as misunderstandings in the AI condition, while refraining from doing so in the human condition for social reasons. To maximize ecological validity and capture the dynamic nature of real-world communication, future paradigms could address this by investigating live, real-time interaction, utilizing LLMs specifically fine-tuned for irony generation. Additionally, because our study used text-based interaction with LLM-powered AI, lacking embodied cues shown to trigger an intentional stance towards robots[22], it raises a question for future research: Would multimodal AI systems (with visual avatars, prosody, or gestures) elicit greater intentional stance adoption via these social-communicative cues, or would increased human-likeness trigger uncanny valley effects[91] that reduce mental state ascription?

## Conclusion

This study demonstrates that people adopt a partial intentional stance when confronted with AI-attributed irony. Behaviorally, although participants recognized irony from both an AI companion and a human, they were less likely to categorize literally incongruent utterances from AI as deliberate communicative acts and more likely to interpret them as errors resulting from misunderstanding the context. This behavioral pattern was mirrored neurally: attenuated P200 and P600 effects for AI-attributed irony indicated reduced cognitive resources in both early semantic processing and late pragmatic reanalysis needed to recover intended meaning, compared to human-attributed irony. These neural responses were modulated by individual perceptions of AI sincerity and trustworthiness, indicating that intentional stance adoption toward AI operates on a flexible continuum shaped by people's mental models. Consequently, achieving social agency in human-AI interaction requires more than just communicative competence. Despite the advanced linguistic capabilities of current LLMs, their outputs have not yet crossed the threshold required to elicit full intentional attribution. Bridging this gap will require not only technological advancements but also a deeper understanding of how human mental models of artificial agents are formed and adapted.

## Data availability

All data and experimental materials freely available on the Open Science Framework at https://osf.io/uay8p.

## Code availability

The code used to generate the results of this study are freely available on the Open Science Framework at https://osf.io/uay8p.

## References

1. García-Méndez, S., de Arriba-Pérez, F. & Somoza-López, M. del C. A Review on the Use of Large Language Models as Virtual Tutors. *Sci. Educ.* **34**, 877–892 (2025).

2. Peng, Y. *et al.* Scaffolding human and AI instruction: Neural alignment and learning gains in online education. *Neuron* **0**, (2026).

3. Casu, M., Triscari, S., Battiato, S., Guarnera, L. & Caponnetto, P. AI Chatbots for Mental Health: A Scoping Review of Effectiveness, Feasibility, and Applications. *Appl. Sci.* **14**, 5889 (2024).

4. Yhee, Y. & Koo, C. Seeing AI as human or machine? Effects of transparency, valence, and readability on review summary helpfulness. *Tour. Manag.* **113**, 105305 (2026).

5. Maples, B., Cerit, M., Vishwanath, A. & Pea, R. Loneliness and suicide mitigation for students using GPT3-enabled chatbots. *Npj Ment. Health Res.* **3**, 4 (2024).

6. Avetisyan, H., Safikhani, P. & Broneske, D. Laughing Out Loud – Exploring AI-Generated and Human-Generated Humor. SSRN Scholarly Paper at https://papers.ssrn.com/abstract=4707459 (2023).

7. Ritschel, H., Aslan, I., Sedlbauer, D. & André, E. Irony Man: Augmenting a Social Robot with the Ability to Use Irony in Multimodal Communication with Humans. in *Proceedings of the 18th International Conference on Autonomous Agents and MultiAgent Systems* 86–94 (International Foundation for Autonomous Agents and Multiagent Systems, Richland, SC, 2019).

8. Dennett, D. C. *The Intentional Stance*. (MIT Press, 1989).

9. Leslie, A. M. Pretending and believing: issues in the theory of ToMM. *Cognition* **50**, 211–238 (1994).

10. Baron-Cohen, S. *Mindblindness: An Essay on Autism and Theory of Mind*. (MIT Press, 1997).

11. Decety, J. & Jackson, P. L. The Functional Architecture of Human Empathy. *Behav. Cogn. Neurosci. Rev.* **3**, 71–100 (2004).

12. Nass, C. & Moon, Y. Machines and Mindlessness: Social Responses to Computers.

*J. Soc. Issues* **56**, 81–103 (2000).
13. Nass, C., Moon, Y. & Carney, P. Are People Polite to Computers? Responses to Computer-Based Interviewing Systems. *J. Appl. Soc. Psychol.* **29**, 1093–1109 (1999).
14. Fogg, B. & Nass, C. How users reciprocate to computers: an experiment that demonstrates behavior change. in *CHI '97 Extended Abstracts on Human Factors in Computing Systems* 331–332 (Association for Computing Machinery, New York, NY, USA, 1997). doi:10.1145/1120212.1120419.
15. Nass, C., Isbister, K. & Lee, E.-J. Truth is beauty: researching embodied conversational agents. in *Embodied conversational agents* 374–402 (MIT Press, Cambridge, MA, USA, 2001).
16. Finkel, S. E., Guterbock, T. M. & Borg, M. J. Race-of-interviewer effects in a preelection poll virginia 1989. *Public Opin. Q.* **55**, 313–330 (1991).
17. Kane, E. W. & Macaulay, L. J. Interviewer gender and gender attitudes. *Public Opin. Q.* **57**, 1–28 (1993).
18. Epley, N., Waytz, A. & Cacioppo, J. T. On seeing human: A three-factor theory of anthropomorphism. *Psychol. Rev.* **114**, 864–886 (2007).
19. Utsumi, A., Watanabe, Y. & Wakayama, Y. Do people understand irony from computers?
20. Krach, S. *et al.* Can Machines Think? Interaction and Perspective Taking with Robots Investigated via fMRI. *PLOS ONE* **3**, e2597 (2008).
21. Marchesi, S. *et al.* Do We Adopt the Intentional Stance Toward Humanoid Robots? *Front. Psychol.* **10**, (2019).
22. Thellman, S., de Graaf, M. & Ziemke, T. Mental State Attribution to Robots: A Systematic Review of Conceptions, Methods, and Findings. *J Hum-Robot Interact* **11**, 41:1-41:51 (2022).
23. Cai, Z., Duan, X., Haslett, D., Wang, S. & Pickering, M. Do large language models resemble humans in language use? in *Proceedings of the Workshop on Cognitive Modeling and Computational Linguistics* (eds Kuribayashi, T., Rambelli, G., Takmaz, E., Wicke, P. & Oseki, Y.) 37–56 (Association for Computational Linguistics, Bangkok,

Thailand, 2024). doi:10.18653/v1/2024.cmcl-1.4.

24. Wei, Z. *et al.* Implicit Perception of Differences between NLP-Produced and Human-Produced Language in the Mentalizing Network. *Adv. Sci.* **10**, 2203990 (2023).

25. Saxe, R., Carey, S. & Kanwisher, N. Understanding Other Minds: Linking Developmental Psychology and Functional Neuroimaging. *Annu. Rev. Psychol.* **55**, 87–124 (2004).

26. Frith, U. & Frith, C. D. Development and neurophysiology of mentalizing. *Philos. Trans. R. Soc. B Biol. Sci.* **358**, 459–473 (2003).

27. Cai, Z. G. Interlocutor modelling in comprehending speech from interleaved interlocutors of different dialectic backgrounds. *Psychon. Bull. Rev.* **29**, 1026–1034 (2022).

28. Zhang, Y. & Cai, Z. G. Perspective alignment in dialogue: The role of interlocutor's linguistic competence and communicative goals. *Cognition* **273**, 106526 (2026).

29. Cai, Z. G., Sun, Z. & Zhao, N. Interlocutor modelling in lexical alignment: The role of linguistic competence. *J. Mem. Lang.* **121**, 104278 (2021).

30. Wu, H. & Cai, Z. G. Speaker effects in language comprehension: An integrative model of language and speaker processing. *Psychon. Bull. Rev.* **33**, 138 (2026).

31. Wu, H., Duan, X. & Cai, Z. G. Speaker Demographics Modulate Listeners' Neural Correlates of Spoken Word Processing. *J. Cogn. Neurosci.* **36**, 2208–2226 (2024).

32. Wu, H., Rao, X. & Cai, Z. G. Probabilistic adaptation of language comprehension for individual speakers: evidence from neural oscillations. *Soc. Cogn. Affect. Neurosci.* **20**, nsaf085 (2025).

33. Wu, H. & Cai, Z. G. When a Man Says He Is Pregnant: Event-related Potential Evidence for a Rational Account of Speaker-contextualized Language Comprehension. *J. Cogn. Neurosci.* **38**, 545–560 (2026).

34. Cai, Z. G. *et al.* Accent modulates access to word meaning: Evidence for a speaker-model account of spoken word recognition. *Cognit. Psychol.* **98**, 73–101 (2017).

35. Rao, X., Wu, H. & Cai, Z. G. Comprehending semantic and syntactic anomalies in text attributed to an LLM versus a human: An ERP study. *Appl. Psycholinguist.* **46**, e51

(2025).

36. Kutas, M. & Federmeier, K. D. Thirty Years and Counting: Finding Meaning in the N400 Component of the Event-Related Brain Potential (ERP). *Annu. Rev. Psychol.* **62**, 621–647 (2011).

37. Hagoort, P. & Brown, C. M. ERP effects of listening to speech compared to reading: the P600/SPS to syntactic violations in spoken sentences and rapid serial visual presentation. *Neuropsychologia* **38**, 1531–1549 (2000).

38. Rao, X., Wu, H. & Cai, Z. G. A funny companion: Distinct neural responses to perceived AI- versus human-generated humor. Preprint at https://doi.org/10.48550/arXiv.2509.10847 (2025).

39. Chang, Y.-T., Ku, L.-C. & Chen, H.-C. Sex differences in humor processing: An event-related potential study. *Brain Cogn.* **120**, 34–42 (2018).

40. Ferrari, V. *et al.* Novelty and emotion: Pupillary and cortical responses during viewing of natural scenes. *Biol. Psychol.* **113**, 75–82 (2016).

41. Sperber, D. Irony and the use-mention distinction. *Radic. Pragmat.* 295–318 (1981).

42. Beres, A. M. Time is of the Essence: A Review of Electroencephalography (EEG) and Event-Related Brain Potentials (ERPs) in Language Research. *Appl. Psychophysiol. Biofeedback* **42**, 247–255 (2017).

43. Regel, S., Gunter, T. C. & Friederici, A. D. Isn't It Ironic? An Electrophysiological Exploration of Figurative Language Processing. *J. Cogn. Neurosci.* **23**, 277–293 (2011).

44. Regel, S., Meyer, L. & Gunter, T. C. Distinguishing Neurocognitive Processes Reflected by P600 Effects: Evidence from ERPs and Neural Oscillations. *PLOS ONE* **9**, e96840 (2014).

45. Regel, S. & Gunter, T. C. Don't Get Me Wrong: ERP Evidence from Cueing Communicative Intentions. *Front. Psychol.* **8**, (2017).

46. Regel, S., Coulson, S. & Gunter, T. C. The communicative style of a speaker can affect language comprehension? ERP evidence from the comprehension of irony. *Brain Res.* **1311**, 121–135 (2010).

47. Landi, N. & Perfetti, C. A. An electrophysiological investigation of semantic and phonological processing in skilled and less-skilled comprehenders. *Brain Lang.* **102**, 30–45 (2007).

48. Wlotko, E. W. & Federmeier, K. D. Finding the right word: Hemispheric asymmetries in the use of sentence context information. *Neuropsychologia* **45**, 3001–3014 (2007).

49. Donchin, E. & Coles, M. G. H. Is the P300 component a manifestation of context updating? *Behav. Brain Sci.* **11**, 357–374 (1988).

50. Ruchkin, D. S., Johnson Jr., R., Canoune, H. L., Ritter, W. & Hammer, M. Multiple Sources of P3b Associated with Different Types of Information. *Psychophysiology* **27**, 157–176 (1990).

51. Gouvea, A. C., Phillips, C., Kazanina, N. & Poeppel, D. The linguistic processes underlying the P600. *Lang. Cogn. Process.* **25**, 149–188 (2010).

52. van Herten, M., Kolk, H. H. J. & Chwilla, D. J. An ERP study of P600 effects elicited by semantic anomalies. *Cogn. Brain Res.* **22**, 241–255 (2005).

53. Nieuwland, M. S. & Van Berkum, J. J. A. Testing the limits of the semantic illusion phenomenon: ERPs reveal temporary semantic change deafness in discourse comprehension. *Cogn. Brain Res.* **24**, 691–701 (2005).

54. Spotorno, N., Cheylus, A., Henst, J.-B. V. D. & Noveck, I. A. What's behind a P600? Integration Operations during Irony Processing. *PLOS ONE* **8**, e66839 (2013).

55. Zhou, P., Garnsey, S. & Christianson, K. Is imagining a voice like listening to it? Evidence from ERPs. *Cognition* **182**, 227–241 (2019).

56. Caffarra, S., Wolpert, M., Scarinci, D. & Mancini, S. Who are you talking to? The role of addressee identity in utterance comprehension. *Psychophysiology* **57**, e13527 (2020).

57. Foucart, A., Santamaría-García, H. & Hartsuiker, R. J. Short exposure to a foreign accent impacts subsequent cognitive processes. *Neuropsychologia* **129**, 1–9 (2019).

58. Liu, B. & Sundar, S. S. Should Machines Express Sympathy and Empathy? Experiments with a Health Advice Chatbot. *Cyberpsychology Behav. Soc. Netw.* **21**,

625–636 (2018).

59. Rheu, M. (MJ), Dai, Y. (Nancy), Meng, J. & Peng, W. When a Chatbot Disappoints You: Expectancy Violation in Human-Chatbot Interaction in a Social Support Context. *Commun. Res.* **51**, 782–814 (2024).

60. Dong, Y. & Wu, Y. Interacting with Healthcare Chatbot: Effects of Status Cues and Message Contingency on AI Credibility Assessment. *Int. J. Human–Computer Interact.* **41**, 6908–6920 (2025).

61. McAllister, D. J. Affect- and Cognition-Based Trust as Foundations for Interpersonal Cooperation in Organizations. *Acad. Manage. J.* **38**, 24–59 (1995).

62. Weng, X., Jiang, X. & Liao, Q. Using information of relationship closeness in the comprehension of Chinese ironic criticism: Evidence from behavioral experiments. *J. Pragmat.* **215**, 55–69 (2023).

63. Chen, X., Li, D. & Wang, X. Uighur college students' irony comprehension in Chinese. *Int. J. Biling.* **26**, 450–475 (2022).

64. Kelley, J. F. An iterative design methodology for user-friendly natural language office information applications. *ACM Trans Inf Syst* **2**, 26–41 (1984).

65. Rheu, M. (MJ), Dai, Y. (Nancy), Meng, J. & Peng, W. When a Chatbot Disappoints You: Expectancy Violation in Human-Chatbot Interaction in a Social Support Context. *Commun. Res.* **51**, 782–814 (2024).

66. Leggitt, J. S. & Gibbs, R. W. Emotional Reactions to Verbal Irony. *Discourse Process.* **29**, 1–24 (2000).

67. Oostenveld, R., Fries, P., Maris, E. & Schoffelen, J.-M. FieldTrip: Open Source Software for Advanced Analysis of MEG, EEG, and Invasive Electrophysiological Data. *Comput. Intell. Neurosci.* **2011**, 156869 (2011).

68. Luck, S. J. *An Introduction to the Event-Related Potential Technique, Second Edition*. (MIT Press, 2014).

69. Winkler, I., Haufe, S. & Tangermann, M. Automatic Classification of Artifactual ICA-Components for Artifact Removal in EEG Signals. *Behav. Brain Funct.* **7**, 30 (2011).

70. Ghandeharion, H. & Erfanian, A. A fully automatic ocular artifact suppression from EEG data using higher order statistics: Improved performance by wavelet analysis. *Med. Eng. Phys.* **32**, 720 (2010).

71. Matuschek, H., Kliegl, R., Vasishth, S., Baayen, H. & Bates, D. Balancing Type I error and power in linear mixed models. *J. Mem. Lang.* **94**, 305–315 (2017).

72. Luck, S. J. & Gaspelin, N. How to get statistically significant effects in any ERP experiment (and why you shouldn't). *Psychophysiology* **54**, 146–157 (2017).

73. Martin, C. D., Garcia, X., Potter, D., Melinger, A. & Costa, A. Holiday or vacation? The processing of variation in vocabulary across dialects. *Lang. Cogn. Neurosci.* **31**, 375–390 (2016).

74. Martin, C. D., Molnar, M. & Carreiras, M. The proactive bilingual brain: Using interlocutor identity to generate predictions for language processing. *Sci. Rep.* **6**, 26171 (2016).

75. Wu, H. & Cai, Z. G. When a Man Says He Is Pregnant: Event-related Potential Evidence for a Rational Account of Speaker-contextualized Language Comprehension. https://doi.org/10.1162/JOCN.a.102 doi:10.1162/JOCN.a.102.

76. Ryskin, R. *et al.* An ERP index of real-time error correction within a noisy-channel framework of human communication. *Neuropsychologia* **158**, 107855 (2021).

77. Luck, S. J. & Gaspelin, N. How to get statistically significant effects in any ERP experiment (and why you shouldn't). *Psychophysiology* **54**, 146–157 (2017).

78. Chen, X. & Zhang, C. Setting the "tone" first and then integrating it into the syllable: An EEG investigation of the time course of lexical tone and syllable encoding in Mandarin word production. *J. Mem. Lang.* **140**, 104575 (2025).

79. Li, X., Ren, G., Zheng, Y. & Chen, Y. How does dialectal experience modulate anticipatory speech processing? *J. Mem. Lang.* **115**, 104169 (2020).

80. Becker, B. Will our social brain inherently shape and be shaped by interactions with AI? *Neuron* **113**, 2037–2041 (2025).

81. Dou, X., Wu, C.-F., Lin, K.-C., Gan, S. & Tseng, T.-M. Effects of Different Types of Social Robot Voices on Affective Evaluations in Different Application Fields. *Int. J.*

*Soc. Robot.* **13**, 615–628 (2021).

82. Powers, A. & Kiesler, S. The advisor robot: tracing people's mental model from a robot's physical attributes. in *Proceedings of the 1st ACM SIGCHI/SIGART conference on Human-robot interaction* 218–225 (Association for Computing Machinery, New York, NY, USA, 2006). doi:10.1145/1121241.1121280.

83. Loy, J. E. & Demberg, V. Perspective Taking Reflects Beliefs About Partner Sophistication: Modern Computer Partners Versus Basic Computer and Human Partners. *Cogn. Sci.* **47**, e13385 (2023).

84. Dunn, M. S. & Cai, Z. G. Linguistic alignment of redundancy usage in human-human and human-computer interaction. *Appl. Psycholinguist.* **46**, e22 (2025).

85. Shevlin, H. Three frameworks for AI mentality. *Front. Psychol.* **17**, (2026).

86. de Lima Prestes, J. A. Pseudo-Consciousness in Artificial Intelligence: a functional and governance framework for consciousness-like systems. https://philsci-archive.pitt.edu/28958/ (2026).

87. Baron-Cohen, S., Wheelwright, S., Hill, J., Raste, Y. & Plumb, I. The "Reading the Mind in the Eyes" Test Revised Version: A Study with Normal Adults, and Adults with Asperger Syndrome or High-functioning Autism. *J. Child Psychol. Psychiatry* **42**, 241–251 (2001).

88. Van Berkum, J. J. A., van den Brink, D., Tesink, C. M. J. Y., Kos, M. & Hagoort, P. The Neural Integration of Speaker and Message. *J. Cogn. Neurosci.* **20**, 580–591 (2008).

89. Lee, J. D. & See, K. A. Trust in Automation: Designing for Appropriate Reliance. *Hum. Factors* **46**, 50–80 (2004).

90. Gambino, A., Fox, J. & Ratan, R. A. Building a stronger CASA: Extending the computers are social actors paradigm. *Hum.-Mach. Commun.* **1**, 71–85 (2020).

91. Mori, M., MacDorman, K. F. & Kageki, N. The Uncanny Valley [From the Field]. *IEEE Robot. Autom. Mag.* **19**, 98–100 (2012).

**Author contributions**

Xiaohui Rao: Conceptualization, Methodology, Formal analysis, Investigation, Visualization, Writing-Original Draft, Writing-Review & Editing. Hanlin Wu: Conceptualization, Methodology, Formal analysis, Investigation. Zhenguang G. Cai: Conceptualization, Writing-Review & Editing, Supervision, Funding acquisition.

**Funding**

This work was supported by General Research Fund (14601525), University Grants Committee, Hong Kong, Do people treat personalized large language models as humans in communication? The funding agency played no role in the study design, data collection, analysis and interpretation of data, or the writing of this manuscript.

**Ethics declarations**

*Competing interest*

The authors declare no competing interests.

**Figure 1**. Flowchart of the irony recognition and categorization task.

**Figure 2**. Study experimental design.

Flowchart of the experimental procedure. Participants were assigned to an AI or a human condition, completed the main interaction task, followed by a manipulation check and an irony recognition and categorization task. General perceptions of AI were collected at the beginning (AI condition) and at the end (human condition) of the study.

**Figure 3**. Distribution of incongruent responses.

The distribution of reasons for incongruent responses across Attribution (human vs AI). Solid points represent the group-level mean of participant proportions for each category. Error bars represent 95% confidence intervals of the mean. Semi-transparent jittered points display the underlying distribution of individual participant-level proportions ($N$ = 59 participants; $N$ = 30 in the human condition, $N$ = 29 in the AI condition).

**Figure 4.** Neural responses to ironic and literal statements.

(A) Amplitudes of ironic and literal statements in the AI and human conditions; (B) Topographies of the Irony effects (Ironic minus Control) during 150-300 ms, 300-450 ms, and 450-900 ms after critical word onset, and spatial visualization of the electrode cluster of 62 central channels for amplitude analyses (see Table S1). All data presented in this figure are derived from $N$ = 59 participants ($N$ = 30 in the human condition, $N$ = 29 in the AI condition).

**Figure 5**. Modulation of irony processing by individual perceptions.

(A) Estimated marginal effects of perceived AI sincerity on the P200 (150-300 ms) and P600 (450-900 ms); (B) Estimated marginal effects of perceived AI trustworthiness on the P300 (300-450 ms) and P600 (450-900 ms). Solid lines represent model-predicted ERP amplitudes for ironic and literal conditions as a function of standardized ratings of perceived AI sincerity and trustworthiness. Shaded regions indicate 95% confidence intervals derived from the LME models ($N$ = 29 participants in the AI condition). The vertical dashed line at Z = 0 denotes the mean rating.